# "Predictive Simultaneous Interpretation: Harnessing Large Language Models for Democratizing Real-Time Multilingual Communication"

**Kurando Iida, Kenjiro Mimura, Nobuo Ito**
ErudAite Inc.
Tokyo, Japan

**Abstract:**
This study introduces a groundbreaking approach to simultaneous interpretation by directly leveraging the predictive capabilities of Large Language Models (LLMs). We present a novel algorithm that generates real-time translations by predicting speaker utterances and expanding multiple possibilities in a tree-like structure. This method demonstrates unprecedented flexibility and adaptability, potentially overcoming the structural differences between languages more effectively than existing systems. Our theoretical analysis, supported by illustrative examples, suggests that this approach could lead to more natural and fluent translations with minimal latency. The primary purpose of this paper is to share this innovative concept with the academic community, stimulating further research and development in this field. We discuss the theoretical foundations, potential advantages, and implementation challenges of this technique, positioning it as a significant step towards democratizing multilingual communication.

# 1 Introduction:

Simultaneous interpretation, the real-time translation of spoken language, is a crucial tool in our increasingly globalized world. Despite significant advancements in artificial intelligence and natural language processing, current automated simultaneous interpretation systems still struggle to match the flexibility and adaptability of human interpreters. This paper proposes a paradigm shift in approach, directly utilizing the predictive capabilities of Large Language Models (LLMs) to create a more dynamic and context-aware simultaneous interpretation system.

Recent developments in LLMs, such as GPT-3 (Brown et al., 2020), have demonstrated remarkable abilities in language understanding and generation. However, their application in simultaneous interpretation has been limited, primarily using them as translation engines rather than leveraging their predictive capabilities. Our research builds upon and significantly extends the work of Zheng et al. (2020), who explored the use of GPT-2 for simultaneous translation.

We propose an algorithm that not only predicts the speaker's next words but also generates and maintains multiple possible translation paths simultaneously. This approach allows for rapid adaptation to changes in speaker intent or subject matter, a capability that has been a significant challenge for existing systems (Ma et al., 2019).

The primary goal of this paper is to share this foundational idea with the broader academic and research community. While ErudAite, the company behind this research, is actively working on implementing this technology with plans for service launch by 2026, we believe that sharing this concept will accelerate progress in the field and contribute to the democratization of simultaneous interpretation.

# 2 Methodology:

Our proposed algorithm consists of five key steps:

**1. Context Acquisition:** Pre-loading the LLM with relevant contextual information to enhance prediction accuracy.

**2. Real-time Transcription:** Converting the speaker's speech into text using advanced speech recognition techniques.

**3. Next Word Prediction and Tree Construction:** Using the LLM to predict multiple possible continuations of the speaker's utterance, forming a tree-like structure of potential translations.

**4. Partial Translation Confirmation:** Confirming and storing partial translations in the target language as soon as they can be determined with high confidence.

**5. Comparison and Output:** Comparing predictions with actual speech and outputting confirmed translations in the target language.

This methodology expands on the incremental translation concept introduced by Oda et al. (2014) and the diverse translation candidate generation proposed by Xu and Carpuat (2021). However, our approach is novel in its application of these concepts in a real-time, LLM-driven system.

# 3 Detailed Explanation of Key Steps:

To illustrate the core functionality of our system, we will focus on Steps 3, 4, and 5, using a representative example:

**Step 3: Next Word Prediction and Tree Construction**

In this step, the system generates a prediction tree based on the current input. Consider the following Japanese input:
「私は昨日、友達と...」 (Watashi wa kinō, tomodachi to...)

The system might generate the following prediction tree:
```
- 「映画を見に行った」(Eiga wo mi ni itta) - "went to see a movie" (Probability: 40%)
- 「食事をした」(Shokuji wo shita) - "had a meal" (Probability: 30%)
```

- 「公園に行った」(Kōen ni itta) - "went to the park" (Probability: 20%)
- Other possibilities (Probability: 10%)

**Step 4: Partial Translation Confirmation**

Based on the input and predictions, the system can confidently translate the beginning of the sentence:
Confirmed partial translation: "Yesterday, I ... with my friend"

The system holds this partial translation in memory while waiting for the sentence to complete.

**Step 5: Comparison and Output**

Let's assume the actual complete utterance is:
「私は昨日、友達と買い物に行った」(Watashi wa kinō, tomodachi to kaimono ni itta)
"Yesterday, I went shopping with my friend"

**The system compares this with the predictions:**
1. None of the high-probability predictions match exactly.
2. The actual ending ("went shopping") falls under the "Other possibilities" category.
3. The system adapts by selecting the "Other" path and generating the appropriate translation.

**Final output: "Yesterday, I went shopping with my friend."**

This example demonstrates the system's ability to:
1. Generate partial translations quickly for improved real-time performance.
2. Handle unexpected utterances by maintaining multiple prediction paths.
3. Adapt swiftly when the actual speech diverges from high-probability predictions.

# 4 Theoretical Analysis:

The core innovation of our approach lies in its direct utilization of LLM's predictive capabilities. By maintaining multiple prediction paths simultaneously, our system can achieve a level of flexibility and adaptability previously unseen in automated interpretation systems. This method allows for:

1. Rapid adaptation to changes in speaker direction or subject matter.
2. More natural handling of language-specific structures and idioms.
3. Potential reduction in interpretation latency without sacrificing accuracy.

Our approach builds upon the wait-time optimization techniques of Ma et al. (2019) but adds a new dimension of flexibility through the use of predictive tree structures.

# 5 Implementation Challenges and Potential Solutions:

While detailed technical specifications are beyond the scope of this paper, we anticipate several key challenges in implementing this system:

**1. Real-time Processing:** Optimizing LLM inference speed for instantaneous predictions.
   Potential Solution: Utilizing specialized hardware and model quantization techniques.

**2. Memory Management:** Efficiently storing and pruning the prediction tree.
   Potential Solution: Implementing adaptive pruning algorithms based on prediction confidence.

**3. Error Detection and Recovery:**
   a) Mismatched Predictions: When the actual speech diverges significantly from all predicted paths.
      Solution: Implement a confidence threshold. If all predictions fall below this threshold, trigger a rapid re-prediction process using the most recent context.

   b) Contextual Errors: When the system misinterprets the overall context of the conversation.
      Solution: Periodically re-evaluate the broader context using a separate LLM module. If a contextual shift is detected, update the main translation model accordingly.

   c) Language-Specific Errors: Mistakes arising from idiomatic expressions or culture-specific references.
      Solution: Incorporate a database of common idioms and cultural references, flagging potential misinterpretations for special handling.

   d) Technical Failures: Such as temporary loss of audio input or system lag.
      Solution: Implement a robust buffering system that can temporarily store audio input. In case of system lag, provide a summarized catch-up translation once the system recovers.

**4. Multilingual Adaptability:** Ensuring consistent performance across diverse language pairs.
   Potential Solution: Fine-tuning LLMs on multilingual datasets and implementing language-specific post-processing.

# 6 Societal Impact and Ethical Considerations:

The democratization of simultaneous interpretation through this technology has the potential to revolutionize global communication. It could significantly impact fields such as:

1. Education: Enabling real-time translation in international classrooms and conferences.
2. Diplomacy: Facilitating more direct communication in multilateral negotiations.
3. Healthcare: Improving doctor-patient communication in diverse communities.
4. Business: Enhancing international collaboration and breaking down language barriers in global markets.
5. Emergency Response: Facilitating communication during international disaster relief efforts.

While this technology aims to make simultaneous interpretation more accessible, we acknowledge potential challenges such as privacy concerns and data security. We are committed to addressing these issues through robust data protection measures and transparent use policies.

# 7 Future Research Directions:

We encourage the academic community to build upon this foundational idea. Potential areas for future research include:

1. Applying the algorithm to diverse language pairs, especially those with significant structural differences.
2. Adapting the system for specific domains such as medical or legal interpretation.
3. Integrating multimodal information (e.g., gestures, facial expressions) to enhance prediction accuracy.
4. Developing comprehensive evaluation metrics for simultaneous interpretation systems.
5. Exploring the cognitive implications of AI-assisted interpretation on human language processing.
6. Extending the predictive approach to real-time natural conversation: This technology can be adapted for generating responsive dialogue in real-time interactions, potentially reducing latency in AI-assisted conversations.

# 8 Broader Applications:

While this paper primarily focuses on simultaneous interpretation, the proposed predictive approach has potential applications beyond translation. By adapting the algorithm to generate responses instead of translations, this technology could revolutionize real-time, natural conversations with AI systems. In this application, the system would predict potential user utterances and pre-generate appropriate responses. This predictive response generation could significantly reduce latency in AI-assisted conversations, making interactions more natural and fluid. Such an approach could have profound implications for various fields, including:

Customer Service: Enabling more responsive and natural chatbots and virtual assistants.
Educational Technology: Facilitating more engaging and interactive AI tutors.
Accessibility: Providing real-time communication assistance for individuals with speech or hearing impairments.
Entertainment: Enhancing NPCs (Non-Player Characters) in video games for more realistic and dynamic interactions.

Further research is needed to adapt and optimize the proposed algorithm for these broader applications, but the potential for improving human-AI interaction across multiple domains is significant.

# 9 Conclusion:

This paper presents a novel approach to simultaneous interpretation that harnesses the full potential of Large Language Models. By leveraging LLMs' predictive capabilities in a more direct and comprehensive manner than previous research, our method has the potential to significantly advance the field of automated simultaneous interpretation and democratize multilingual communication.

While substantial challenges remain in implementation and evaluation, this approach opens new avenues for research and development in natural language processing and machine translation. As LLM technology continues to evolve, we anticipate that this predictive simultaneous interpretation paradigm will play a crucial role in breaking down language barriers and facilitating global communication.

We call for collaboration between academia and industry to further develop and refine this technology. By working together, we can accelerate progress towards making real-time, high-quality simultaneous interpretation accessible to all, fostering greater understanding and cooperation in our increasingly interconnected world.

# Acknowledgements:

This research was conceptualized and rigorously developed by Kurando Iida, CEO of ErudAite with Kenjiro Mimura and Nobuo Ito. The writing process was assisted by the artificial intelligence system Claude 3.5 Sonnet. Claude 3.5 Sonnet, under Iida's guidance, organized his original ideas and meticulous analysis into an academic paper format, expressing them in line with scholarly conventions.

The innovative approach proposed in this study is based on Iida's deep insights and creativity, opening new possibilities in simultaneous interpretation technology. Claude 3.5 Sonnet's contribution was instrumental in systematically documenting these ideas and clarifying their relevance to existing academic research.

This collaboration demonstrates the potential for more effective dissemination of research outcomes by combining human creativity with AI capabilities.

The author expresses sincere gratitude to all those involved in this research and to the broader scientific community for their continued efforts in advancing the field of natural language processing and machine translation.